\documentclass[acmsmall]{acmart}
\AtBeginDocument{%
  }
\usepackage{amsmath,amsfonts}
\usepackage{subfig}
\usepackage{bbm}
\usepackage{stfloats}
\usepackage{multirow}
\usepackage{xspace}
\makeatletter
\DeclareRobustCommand\onedot{\futurelet\@let@token\@onedot}
\def\@onedot{\ifx\@let@token.\else.\null\fi\xspace}
\def\eg{\emph{e.g}\onedot} 
\def\ie{\emph{i.e}\onedot}
\def\etal{\emph{et al}\onedot}
\def\vs{\emph{vs}\onedot}

\setcopyright{acmcopyright}
\copyrightyear{2018}
\acmYear{2018}
\acmDOI{XXXXXXX.XXXXXXX}

\acmJournal{JACM}
\acmVolume{37}
\acmNumber{4}
\acmArticle{111}
\acmMonth{8}

\begin{document}

\title{Noise-Tolerant Hybrid Prototypical Learning with Noisy Web Data}


\author{Chao Liang}
\email{cs.chaoliang@zju.edu.cn}
\affiliation{
  \institution{Zhejiang University}
  \state{Zhejiang}
  \country{China}
  \postcode{310027}
}
\author{Linchao Zhu}
\email{zhulinchao@zju.edu.cn}
\affiliation{
  \institution{Zhejiang University}
  \state{Zhejiang}
  \country{China}
  \postcode{310027}
}
\author{Zongxin Yang}
\email{yangzongxin@zju.edu.cn}
\affiliation{
  \institution{Zhejiang University}
  \state{Zhejiang}
  \country{China}
  \postcode{310027}
}
\author{Wei Chen}
\email{chenvis@zju.edu.cn}
\affiliation{
  \institution{Zhejiang University}
  \state{Zhejiang}
  \country{China}
  \postcode{310027}
}
\author{Yi Yang}
\email{yangyics@zju.edu.cn}
\affiliation{
  \institution{Zhejiang University}
  \state{Zhejiang}
  \country{China}
  \postcode{310027}
}

\renewcommand{\shortauthors}{Liang et al.}

\begin{abstract}
We focus on the challenging problem of learning an unbiased classifier from a large number of potentially relevant but noisily labeled web images given only a few clean labeled images. This problem is particularly practical, because it reduces the expensive annotation costs by utilizing freely accessible web images with noisy labels.
Typically, prototypes are representative images or features used to classify or identify other images.
However, in the few clean and many noisy scenarios, the class prototype can be severely biased due to the presence of irrelevant noisy images. The resulting prototypes are less compact and discriminative, as previous methods do not take into account the diverse range of images in the noisy web image collections. On the other hand, the relation modeling between noisy and clean images is not learned for the class prototype generation in an end-to-end manner, which results in a suboptimal class prototype.
In this paper, we introduce a similarity maximization loss named the SimNoiPro. Our SimNoiPro first generates noise-tolerant hybrid prototypes composed of clean and noise-tolerant prototypes, and then pulls them closer to each other. Our approach considers the diversity of noisy images by explicit division and overcomes the optimization discrepancy issue. This enables better relation modeling between clean and noisy images and helps extract judicious information from the noisy image set.
The evaluation results on two extended few-shot classification benchmarks confirm that our SimNoiPro outperforms prior methods in measuring image relations and cleaning noisy data.
\end{abstract}

\ccsdesc[300]{Computing methodologies~Visual content-based indexing and retrieval}
\ccsdesc[300]{Computing methodologies~Object recognition}

\keywords{Deep learning, learn from noisy labels, few-shot learning, prototypical learning}


\maketitle

\section{Introduction}
Deep learning has revolutionized many computer vision tasks such as classification~\cite{he2016deep, luo2015image, krizhevsky2012imagenet, zhao2024rlcf}, segmentation~\cite{he2017mask, yang2021collaborative, yang2024scalable}, image/video understanding~\cite{zhu2021temporal, fan2020ran, wang2023align, yang2024doraemongpt, zhang2024paa, li2024cir}, object detection~\cite{faster_rcnn, redmon2016you, xu2021training}, and few-shot learning~\cite{Douze_2018_CVPR, finn2017model, Gidaris_2018_CVPR,iscen2020graph, siam2019amp,NIPS2017_cb8da676, NIPS2016_90e13578, deepemd, he_tomm22, shi_tomm, ravi2016optimization, rusu2018meta, he2020memory, santoro2016one}. State-of-the-art visual system is often exceptionally data-hungry and requires massive well-annotated data. To mitigate the demand for labeled data, few-shot visual recognition tends to learn a model that can be easily generalized with only a few labeled examples on the end task. Though the representation is learned with available many labeled images, the few-shot classifier is easily biased with limited examples. 
\begin{figure*}[t]
\center
\includegraphics[width=1.0\linewidth]{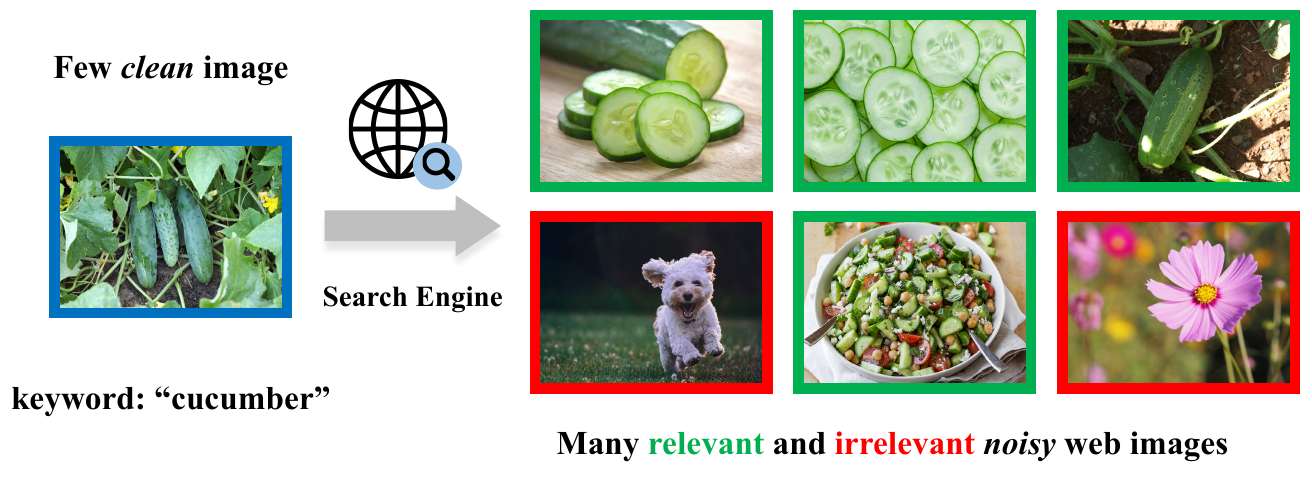}
\caption{
\textbf{Relevant images exist in web images.} Massive freely accessible web images can be obtained by search engines. These images are easily acquired but can be inevitably annotated with noisy labels.
}
\label{fig:intro}  
\end{figure*}

As computational power continues to develop and web data scales up, some researchers have proposed to use large-scale external data sources to enhance the few-shot classifier~\cite{Douze_2018_CVPR, iscen2020graph}. Often, there are a few closely related images among a large number of irrelevant noisy images~(Figure~\ref{fig:intro}). Prototypical learning is widely used for classification~\cite{NIPS2017_cb8da676, li2021mopro} where the class prototypes are the representative images or features used to classify or identify other images. However, the presence of irrelevant noisy images can severely affect the generation of robust class prototypes. To improve the generalization of such few-shot learners, the existing works have attempted to identify the relevance between noisy images and the support clean image set and assigned high relevance scores to those relevant images in the noisy image set. Once relevance is accurately established, those relevant images in the noisy set could be beneficial for learning the noise-tolerant class prototypes.

Recently, Iscen~\etal~\cite{iscen2020graph} proposed to use a graph convolutional neural network~(GCN) to address the problem of few clean and many noisy examples. The GCN is optimized using a binary classifier to discriminate clean from noisy images. By modeling with the graph structure, this method considers the relationships between clean and noisy examples, generating a relevance score for each noisy example based on its visual similarity to clean examples. The results of this GCN-based approach are very promising, demonstrating that noisy examples can benefit few-shot classifier learning.
However, the approach neglects the considerable diversity within the noisy image set. It leverages only one global noise-tolerant prototype to model the noisy set at a coarse level. This approach fails to represent the complex noisy collections of diverse web images effectively. Hence, it might produce unsatisfactory relevance scores.
Furthermore, using the binary cross entropy loss as the training objective results in a discrepancy issue. The binary classifier treats all the web images with noisy labels as negative instances. It forces the model to produce low relevance scores for those relevant images in the noisy set, even though those visually similar examples can contribute to the learning of the few-shot classifier. Also, the relevance scores are not learned for the class prototype generation in an end-to-end manner, which leads to a suboptimal solution.

To address the challenges posed by the diverse noisy image set and the optimization discrepancy issue, we propose a similarity maximization loss named the SimNoiPro, for the few clean and many noisy problem. Our SimNoiPro involves two steps. 

\textbf{First}, we build noise-tolerant hybrid prototypes to model the complex data collections, by dividing the noisy set into a few groups based on their relevance scores. Typically, noisy and clean data hold distinct properties. Noisy data points are more widely dispersed in the feature space. Using one coarse noise-tolerant prototype may significantly bias against the few clean samples and lead to large intra-class variance. The introduction of multiple noise-tolerant prototypes offers more refined modeling of the large-scale diverse noisy set.

\textbf{Second}, we propose a similarity maximization loss to generate more compact and discriminative class prototypes. A good prototype should exhibit the characteristics of compactness and discrimination. This requires pulling the noise-tolerant prototypes and the clean prototype closer to each other, which can produce a tighter cluster and diminish intra-class variance. Also, our similarity maximization loss bridges the optimization between relevance scores and the class prototype generation in an end-to-end manner. It overcomes the optimization discrepancy issue caused by the binary classification loss, resulting in a more plausible and effective clean-noisy relation.

Our proposed approach yields promising results on two standard few-shot classification benchmarks, \ie, Low-shot Places365 and Low-shot ImageNet. Notably, we outperform~\cite{iscen2020graph} by 4.4\% and 3.5\% in the 5-shot setting for Low-shot Places365 and Low-shot ImageNet, respectively.

We summarize our contributions in the following:
\begin{itemize}
    \item We propose noise-tolerant hybrid prototypes to handle the diversity of the noisy web data.
    \item We propose a similarity maximization loss to bridge the optimization between the relevance scores and class prototype generation in an end-to-end manner, generating compact and discriminative class prototypes.
    \item Extensive experiments on Low-shot Places365 and Low-shot ImageNet demonstrate the effectiveness of our proposed method, which outperforms other baselines consistently.
\end{itemize}

The rest of the paper is organized as follows. We first introduce the related works in Section~\ref{sec:related_work}. Section~\ref{sec:our_method} reviews the previous GCN cleaning framework and details our proposed SimNoiPro. Section~\ref{sec:experiment} presents performance evaluation and visualization, followed by the conclusion in Section~\ref{sec:conclusion}.

\section{Related Work}
\label{sec:related_work}
\subsection{Few-shot Learning}

In the past few years, there has been a growing interest in few-shot learning, which aims to mitigate the demand for labeled data by learning from a few labeled training examples. The existing few-shot learning research~\cite{Douze_2018_CVPR, finn2017model, Gidaris_2018_CVPR,iscen2020graph, siam2019amp,NIPS2017_cb8da676, Sung_2018_CVPR,NIPS2016_90e13578, deepemd} can be broadly classified into several categories. One of the promising directions is based on the meta-learning paradigm, also known as ``learn to learn''. This paradigm focuses on capturing generic knowledge during the meta-training stage and then adapting to a completely new task rapidly. Specifically, meta-learning based methods follow the episode paradigm in the few-shot regime. Among these methods, metric-based approaches and optimization-based approaches are the two main streams in the literature. Metric-based methods aim to enhance the discriminative features in the embedding space and learn a good distance function over them. For instance, Matching Network~\cite{NIPS2016_90e13578} leverages the attention mechanism to align the query and support examples. Prototypical Network~\cite{NIPS2017_cb8da676} extends this approach to compare the Euclidean distance between the class representations. Furthermore, RelationNet~\cite{Sung_2018_CVPR} proposes to learn a good metric by using deep networks, while DeepEMD~\cite{deepemd} employs the Earth Mover's distance to compute the structure similarity.
In the optimization-based methods, the key idea is to adjust the optimization algorithm so that the model can learn from few data better. Meta-LSTM\cite{ravi2016optimization} proposes the LSTM-based meta-learner to optimize the classifier. MAML~\cite{finn2017model} aims to find good initial parameters where fast convergence can be achieved given limited training examples.

Another class of few-shot learning approaches is based on transfer learning~\cite{Hariharan_2017_ICCV,Gidaris_2018_CVPR,Douze_2018_CVPR}. These methods leverage the large amount of data from base classes to learn a robust model, which is then finetuned on the few labeled training examples from the novel classes, providing the capability to recognize the novel classes. Hariharan~\etal~\cite{Hariharan_2017_ICCV} propose a generative method for data augmentation. Gidaris~\etal~\cite{Gidaris_2018_CVPR} design an attention-based generator to dynamically predict the weights for the novel classes. Douze~\etal~\cite{Douze_2018_CVPR} formulate the few-shot problem as a semi-supervised setting by introducing large-scale collections of unlabeled images.

\subsection{Learn from Noisy Data}

High-quality human-annotated data are costly and time-consuming to obtain~\cite{zhai_tomm, Ricci_tomm}. Recently, several works have resorted to large-scale web images from social media to facilitate the learning of the deep neural network. These images are easily acquired by search engines but can be inevitably annotated with noisy labels. Deep neural networks are prone to overfit on noisy labels~\cite{zhang2016understanding}. Learning with too many noisy labels can impair the generalization of deep models. The existing noise-resistant methods can be primarily divided into three types: 1) label correction by the predictions from the deep models~\cite{ma2018dimensionality,reed2014training, tanaka2018joint,yi2019probabilistic, chao2023stitchup}. 2) sample selection by filtering noisy instances~\cite{arazo2019unsupervised,sun2022boosting, liang2023combating}. 3) 
sample reweighting by assigning confidence scores for each noisy data~\cite{ren2018learning, shu2019meta, jiang2018mentornet}. Particularly, \cite{ren2018learning} and~\cite{shu2019meta} assume there is a small unbiased and clean validation set. This setting is similar to our work, but we follow the few-shot learning setting where deep models are pretrained on large data from base classes and focus more on classifier learning with limited clean and many noisy data in unseen domains. Both \cite{gjs} and \cite{ncr} consider the consistency of the network prediction for noise-robust learning. GJS~\cite{gjs} encourages consistency around data points. NCR~\cite{ncr} proposes an additional loss to penalize the divergence of the predictions from the neighbors in the feature space.

\subsection{Prototypical Learning}
The prototype can be considered as the representation of a cluster of semantically similar instances. Prototype-based learning has been widely applied in learning from noisy labels~\cite{li2021mopro}, unsupervised learning~\cite{swav, li2021prototypical}, and especially few-shot learning~\cite{NIPS2017_cb8da676, tang_tomm}. Class prototype reflects a simple inductive bias, which brings about well-generalized performance in the limited-data regime.
Considering the fact that the noisy web image set is diverse, using only one global prototype cannot represent the whole noisy set well~\cite{yang2021multiple}.

\section{Method}
\label{sec:our_method}
\subsection{Preliminary}
\subsubsection{Problem statement}
We aim to learn an unbiased few-shot classifier with the aid of additional large-scale web data with noisy labels. Generally, there are two stages in the few-shot classification.

The first stage is the representation learning phase. In this stage, we are given a large clean labeled dataset
$\mathcal{D}_{base}=\{\mathcal{D}_{base}^1, \mathcal{D}_{base}^2, \ldots, \mathcal{D}_{base}^{L}\}$, where $L$ is the number of categories in the base set.
$\mathcal{D}_{base}^{l}$ denotes the set of images for base class $l$. 
$\mathcal{D}_{base}^{l}=\{(x_i^{l}, y_i^{l}) | i = 1,...,n\}$, where $n$ is the number of training examples. The base set $\mathcal{D}_{base}$ is leveraged to learn a strong feature extractor $\Phi:x\mapsto\Phi(x) \in \mathbb{R}^d$. Here, $d$ denotes the dimension of the feature.

In the second stage, we receive a novel dataset $\mathcal{D}_{novel}$ with few clean labeled examples. The goal is to adapt the learned feature extractor to the novel dataset and train a robust classifier given only few clean examples.
We define $\mathcal{D}_{novel}=\{\mathcal{D}_{novel}^1, \mathcal{D}_{novel}^2, \ldots, \mathcal{D}_{novel}^C\}$ and $C$ is the number of classes in the novel set. 
For each novel class $c$, $\mathcal{D}_{novel}^c=\{(x_i^{c}, y_i^{c}) | i = 1,...,k\}$, where $k$ is the number of novel examples and  $k \ll n$, usually called the $k$-shot setting. Note that the base classes in $\mathcal{D}_{base}$ have no overlaps with the novel classes in $\mathcal{D}_{novel}$. 

In this paper, we focus on the second stage of few-shot learning and tackle this challenge with an additional large-scale noisy dataset $\mathcal{D}_{noisy}$, particularly from the Internet. The problem is formulated as learning an unbiased classifier from many freely accessible web images with noisy labels given only few clean labeled images. Combining both clean labeled data and large amounts of noisy ones, we obtain the whole dataset $\widetilde{\mathcal{D}}_{novel}^c=\mathcal{D}_{novel}^c\cup\mathcal{D}_{noisy}^c$ for each novel class $c$.
 Note that $\mathcal{D}_{novel}^c$ is the image set with few clean labeled examples and $\mathcal{D}_{noisy}^c$ consists of a large number of images with noisy labels.
 We expect to identify the relevant images from the noisy set. Training with these relevant images can potentially enhance the generalization of the few-shot classifier.

We extract the features of examples in $\widetilde{\mathcal{D}}_{novel}^c$ using the learned feature extractor $\Phi$.  
The clean and noisy features are represented by the matrix $V^c = [\mathbf{v}_1^c, ..., \mathbf{v}_k^c, ..., \mathbf{v}_{N}^c] \in \mathbb{R}^{d\times N}$, where $\mathbf{v}_i^c = \Phi(x_i^c) \in \mathbb{R}^d$, $x_i^c$ is a training example from $\widetilde{\mathcal{D}}_{novel}^c$ and $N$ is the number of training examples in $\widetilde{\mathcal{D}}_{novel}^c$. We assume the first $k$ features are from the clean set and the remaining are from the noisy set. For convenience, we ignore the superscript $c$ if it can be inferred from the context.

\subsubsection{Review of GCN cleaning framework}
\label{sec:reviewGCN}
Our work extends the GCN cleaning framework~\cite{iscen2020graph} by directly bridging the optimization between the relevance scores and noise-tolerant hybrid prototypes generation in an end-to-end manner.
Firstly, we review the relation modeling with GCN.

Iscen~\etal~\cite{iscen2020graph} introduced a noise cleaning framework to tackle the few clean and many noisy problem. Their approach is divided into two stages: a graph cleaning and a classifier learning stage. 

First, they applied a two-layered GCN to perform offline cleaning by predicting a class relevance score $r$ for each noisy example in the $\mathcal{D}_{noisy}^{c}$. The class relevance scores $\mathbf{r} \in \mathbb{R}^{N}$ are learned independently for each novel class $c$:
\begin{align}
    \mathbf{r} = F_{\Theta}(\widetilde{A},V) = \textit{Sigmoid}(\Theta_{2}^\top[\Theta_{1}^{\top}V\widetilde{A}]_{+}\widetilde{A}),
\label{eq:r_value}
\end{align}
where $\Theta=\{\Theta_1, \Theta_2\}$ is the GCN parameters,
$[\cdot]_{+}=ReLU(\cdot)$ and $\widetilde{A}$ is the normalized affinity matrix. 
The training process is constrained by a binary classification loss $L_{\Theta}$:
\begin{equation}
L_{\Theta} = - \frac{1}{k}\sum_{i=1}^{k}{\log r_i} 
- { \frac{\lambda}{N-k}}\sum_{i=k+1}^{N}{\log(1-r_i)},
    \label{eq:binary_cls}
\end{equation}
where $\lambda$ is a hyperparameter for balancing. This loss aims to classify the clean labeled images as positive examples and treat all the images with noisy labels as negative examples.

In general, a class prototype $\mathbf{p}_c$ for category $c$ is defined as below:
\begin{equation}
    \mathbf{p}_c = \frac{1}{r(\widetilde{\mathcal{D}}_{novel}^c)}\sum_{i=1}^{N}{r_i\mathbf{v}_i},
\label{eq:class_proto}
\end{equation}
where the normalization term $r(\widetilde{\mathcal{D}}_{novel}^c) = \sum_{i=1}^{N}{r_i^c}$. The prototype classifier $\mathbf{P}$ consists of $C$ prototypes, \ie $\mathbf{P} = [\mathbf{p}_1,...,\mathbf{p}_C] \in \mathbb{R}^{d \times C}$.

In prototypical learning, each class would produce a single prototype to serve as a representative feature for discriminative classification. In the few clean and many noisy classification, the unified prototype could be learned from the combination of a clean prototype and a noise-tolerant prototype based on Eq.~\ref{eq:class_proto}:
\begin{align}
    \mathbf{p}_c &= \mathbf{p}_{clean} + \mathbf{p}_{noise},
\label{eq:class_proto_combo}
\end{align}   
where
\begin{align}
    \mathbf{p}_{clean} &= \frac{1}{r(\widetilde{\mathcal{D}}_{novel}^c)} \sum_{i=1}^{k}{\mathbf{v}_i},
    \label{eq:clean_proto}\\
    \mathbf{p}_{noise} &= \frac{1}{r(\widetilde{\mathcal{D}}_{novel}^c)}\sum_{i=k+1}^{N}{r_i\mathbf{v}_i}.
    \label{eq:noise_proto}
\end{align}

Second, a few-shot classifier is learned over the novel classes. Particularly, they used a cosine classifier to minimize the loss function $L(\widetilde{\mathcal{D}}_{novel};\mathbf{W})$:
\small
\begin{equation}
    L(\widetilde{\mathcal{D}}_{novel} ;\mathbf{W}) = -\sum_{c=1}^{C}{\frac{1}{r(\widetilde{\mathcal{D}}_{novel}^{c})}\sum_{i=1}^{|\widetilde{\mathcal{D}}_{novel}^c|}{r_i^c\log(\boldsymbol{\sigma}(s\mathbf{\hat{W}}^\top \mathbf{\hat{v}}_i^c)_{c})}}.
\label{eq:cosine_loss}
\end{equation}
\normalsize
Herein, $\boldsymbol{\sigma}(\cdot)$ is the softmax function, $s$ is the temperature parameter. Both the feature vectors $\mathbf{v} \in \mathbb{R}^d$ and the classifier $\mathbf{W} \in \mathbb{R}^{d \times C}$ are transformed into $\ell_2$-normalized form, denoted as $\mathbf{\hat{v}}$ and $\mathbf{\hat{W}}$.

Although this approach achieved promising results over those methods that only deal with clean data, one of the drawbacks is that noisy data are often diverse and one noise-tolerant prototype cannot represent the whole set. Meanwhile, the binary classification loss introduces a discrepancy between the graph cleaning process and the subsequent prototype-based classifier learning.
There is no guarantee that the learned $\mathbf{p}_{noise}$ using the class relevance score $r$ would produce a meaningful prototype for classification.
\begin{figure*}[t]
\center
\includegraphics[height=0.33\linewidth]{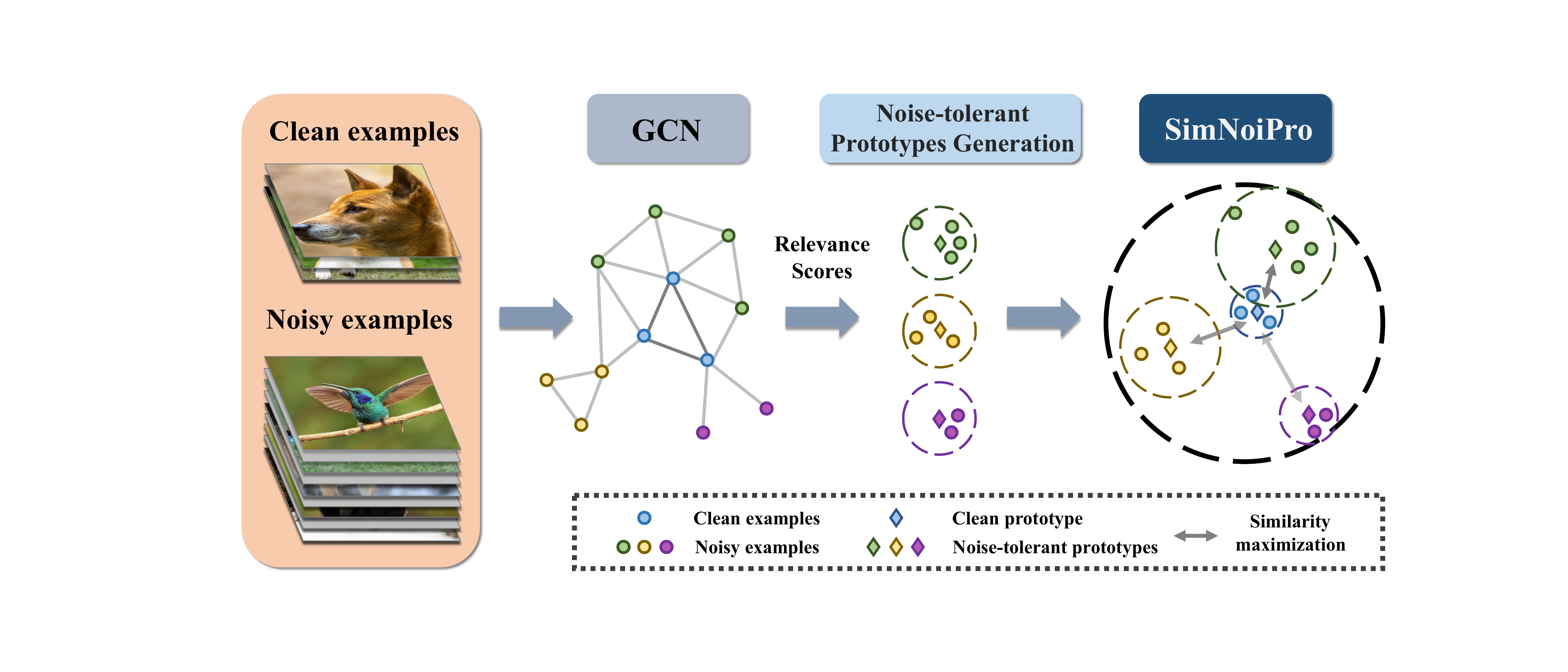}
\caption{
\textbf{Overall framework.} First, SimNoiPro divides noisy examples into several groups by class relevance scores $r$ to produce noise-tolerant prototypes. Then, the similarity maximization loss pulls noise-tolerant prototypes closer to the clean prototype in order to generate a more discriminative prototype for classification.
}
\label{fig:framework}  
\end{figure*}
\subsection{Learning from Noise-tolerant Prototypes}
As discussed in Section~\ref{sec:reviewGCN}, using one global noise-tolerant prototype cannot represent the complex noisy web data, and the binary classification loss (Eq.~\ref{eq:binary_cls}) in~\cite{iscen2020graph} introduces a discrepancy between the training and the inference stage.

We address the obstacles by introducing noise-tolerant hybrid prototypes composed of one clean and multiple noise-tolerant prototypes, and directly optimizing relevance scores and the noise-tolerant prototypes generation ($\mathbf{p}_{noise}$) in an end-to-end manner. \textbf{First}, given the diversity in the noisy set, we divide the noisy examples into a few groups based on their relevance scores and obtain multiple noise-tolerant prototypes (Section~\ref{sec:noisy_proto_gen}).
\textbf{Second}, we propose a similarity maximization loss named the SimNoiPro to pull the noise-tolerant prototypes to be closer to the clean prototype (Section~\ref{sec:simnoisyproto}). The overall framework is presented in Figure~\ref{fig:framework}.

\subsubsection{Noise-tolerant prototypes generation}
\label{sec:noisy_proto_gen}
The diversity of the noisy images motivates us to treat the noisy images differently. In the noisy image set, there might be a few closely relevant images that are visually similar to the clean examples. In the meantime, it is expected that there are a large number of noisy images that are irrelevant. It can be imagined that the noisy features are scattered in the embedding space. If we use one noise-tolerant prototype to represent the noisy set, it can incur a large intra-class variance. Based on this hypothesis, we propose to divide the noisy images into multiple groups and then generate multiple noise-tolerant prototypes. We define the noise-tolerant prototypes generation procedure as $G$. Take the noisy ``feature-relevance score'' pairs as input, $G$ produces $T$ noise-tolerant prototypes based on some criteria,
\begin{equation}
\{\mathbf{p}_{noise}^t\}_{t=1}^T = G(\{(\mathbf{v}_i, r_i)\}_{i=k+1}^{N}).
\end{equation}
Each noise-tolerant prototype $\mathbf{p}_{noise}^t$ is the representative ``cluster'' or ``center'' of the corresponding noise group. Each group also shares similar properties.

Following the above paradigm, we consider two types of noise-tolerant prototypes generation procedure $G$.

\textbf{Feature based clustering.} One of the methods to split the noisy images is to perform clustering on the visual features, \eg k-means, where the noisy images could be grouped into $T$ clusters $\{\mathbf{C}_t\}_{t=1}^T$. The noise-tolerant prototype for each cluster $\mathbf{C}_t$ can be expressed as:
\begin{equation}
    \mathbf{p}_{noise}^t = \sum_{i=k+1}^{N} \mathbbm{1}(\mathbf{v}_i \in \mathbf{C}_t){r_i}\mathbf{v}_i.
\end{equation}
Since the noisy features are pre-computed in our setting, clustering can be done offline before the noise cleaning. One of the disadvantages is that the group is fixed and cannot be further adjusted during the graph cleaning stage.

\textbf{Relevance score based separation.} In this paper, we introduce a general strategy to divide noisy images into multiple groups. We group the noisy images based on the class relevance scores $\mathbf{r}$ (Eq.~\ref{eq:r_value}) generated by the same graph convolutional network $F_{\Theta}(\widetilde{A}, V)$.
For each group $t \in [1, \ldots, T]$, its  relevance score window ${w}_t$ is denoted as:
\begin{equation}
\small
\{r_i | r_{min}+\frac{r_{max}-r_{min}}{T}(t-1) \le r_i < r_{min}+\frac{(r_{max}-r_{min})}{T}{t}\},
\label{eq:noise_window}
\end{equation}
where $r_{max}$ and $r_{min}$ denote the maximum relevance score and the minimum relevance score, respectively.
In each noisy window $w_t$, we define the corresponding noise-tolerant prototype as the weighted average of the features of all noisy images. The noise-tolerant prototype for window $w_t$ (Eq.~\ref{eq:noise_window}) is denoted as:
\begin{equation}
    \mathbf{p}_{noise}^t = \sum_{i=k+1}^{N} \mathbbm{1}{(r_i \in w_t)}{r_i}\mathbf{v}_i.
\end{equation}
The generated noise-tolerant prototype  $\mathbf{p}_{noise}^t$ can be regarded as the representative prototype for all noisy images in window $w_t$. Because the relevance scores are learned from the relations between the noisy images and clean examples, it takes more factors into consideration to produce better separation results. Note that the groups are dynamically changed during the training. We adopt this strategy in most of our experiments.

The introduction of multiple noise-tolerant prototypes provides a separation for different noisy examples, which enables the modeling of the complex noisy web data in a finer manner.

\subsubsection{SimNoiPro Loss}
\label{sec:simnoisyproto}
Given noise-tolerant hybrid prototypes consisted of $T$ diverse noise-tolerant prototypes and one clean prototype $\{\mathbf{p}_{noise}^1, ...,\mathbf{p}_{noise}^t, ..., \mathbf{p}_{noise}^T, \mathbf{p}_{clean}\}$, our objective is to yield a compact and discriminative prototype $\mathbf{p}_c$ for few-shot image classification. The expected prototype is supposed to be represented as the sum of the noise-tolerant prototypes and the clean prototype. However, computed from scarce clean examples and large-scale noisy examples, it is nearly hard to get an unbiased prototype to support the class decision boundary if there are no constraints between the noise-tolerant prototypes and the clean prototype. A discriminative prototype is expected to exhibit low intra-class variance in the few clean and many noisy scenarios. Since the clean examples are more reliable than the noisy ones, it is much more reasonable to require the expected prototype biased towards the clean prototype.

We introduce a similarity maximization loss to pull the noise-tolerant prototypes to be closer to the clean prototype. The cosine distance is used to measure the similarity between two prototypes $\mathbf{p}_1$ and $\mathbf{p}_2$:

\newcommand{\lnorm}[1]{\frac{#1}{\left\lVert{#1}\right\rVert _2}}
\begin{equation}
    M(\mathbf{p}_1, \mathbf{p}_2) =  - \lnorm{\mathbf{p}_1}{\cdot}\lnorm{\mathbf{p}_2},
    \label{eq:sim_max}
\end{equation} 
where ${\left\lVert{\cdot}\right\rVert _2}$ is $\ell_2$-norm. Our goal is to minimize the negative similarity between the two prototypes. The distance minimization enforces the learned noise-tolerant prototype to be relevant to the specific category. It can overcome the discrepancy between the relevance score generation process and the prototype-based classification stage, so as to learn a discriminative unified prototype from the clean prototype and the noise-tolerant prototypes in an end-to-end manner.

It is worth emphasizing that the noise-tolerant prototypes might hold different degrees of bias against the clean prototype. Specifically, we define the SimNoiPro loss based on Eq.~\ref{eq:sim_max} as below:
\begin{equation}
L = \frac{1}{T}\sum_{t=1}^{T} \alpha_t M(\mathbf{p}_{clean}, \mathbf{p}_{noise}^t) + \beta M(\mathbf{p}_{clean}, \mathbf{p}_{noise}),
\end{equation}
where $\mathbf{p}_{clean}$ is the clean prototype in Eq.~\ref{eq:clean_proto}, $\mathbf{p}_{noise}$ is the global noise-tolerant prototype in Eq.~\ref{eq:noise_proto}, $\alpha_t$ is a scaling factor to control the relative weight of the similarity between the clean prototype and each local noise-tolerant prototype and $\beta$ is a hyperparameter. In general, a noise-tolerant prototype that exhibits a higher degree of similarity to the clean prototype should be assigned a relatively more substantial proportion in the loss function.

\textbf{Discussion}. Iscen~\etal~\cite{iscen2020graph} used a binary classification loss during training to classify the noisy examples as negative instances and treat the clean examples as positive instances. It can assign low relevance scores for those relevant images in the noisy set. The relevance scores and the class prototype generation are separated into two stages. No constraints are imposed to maintain the quality of the prototype, which results in a suboptimal class prototype. 
In contrast, our SimNoiPro is easy to implement and it directly optimizes the relevance scores and the class prototype generation in an end-to-end manner, which could potentially produce better relevance scores in a more plausible way. In the experiments, we find SimNoiPro works better in the low-data regime, but comparably in the 1-shot setting. The clean prototype formed from one clean image might incur large intra-class variance, \eg if the clean image is a photo of cucumbers growing on vines, it might be difficult to identify noisy images of cucumber slices in salad as relevant ones.

\section{Experiments}
\label{sec:experiment}
\subsection{Benchmarks and Evaluation}
\subsubsection{Datasets}
We evaluate our method on two benchmarks: Low-shot Places365~\cite{zhou2017places} and Low-shot ImageNet~\cite{Hariharan_2017_ICCV}.
\begin{table*}
\centering
\small
\setlength{\tabcolsep}{3pt}
\caption{
\textbf{Comparisons between different methods on Low-shot Places365.} Our method outperforms other baselines consistently. Iscen~\etal$^\dag$~\cite{iscen2020graph} is reimplemented by ourselves. \textbf{Best} results are highlighted.
}
\begin{tabular}{lccccc}
\toprule
\multirow{2}{*}
{\textbf{Methods}}  & \multicolumn{5}{c}{TOP-5 ACCURACY ON NOVEL CLASSES} \\
 & k=1 & k=2 & k=5 & k=10 & k=20\\ 
\midrule
\multicolumn{6}{c}{ResNet-10 - Few Clean Data} \\
\midrule
Class proto.~\cite{Gidaris_2018_CVPR} & 28.7 $\pm$ 1.12 & 38.0 $\pm$ 0.37 & 50.5 $\pm$ 0.51 & 57.9 $\pm$ 0.35 & 62.3 $\pm$ 0.25 \\
\midrule
\multicolumn{6}{c}{ResNet-10 - Few Clean \& Many Noisy Data} \\
\midrule
$\beta$-weighting, $\beta = 1$~\cite{iscen2020graph} & 44.0 $\pm$ 0.34 & 45.7 $\pm$ 0.22 & 48.4 $\pm$ 0.31 & 50.0 $\pm$ 0.12 & 50.8 $\pm$ 0.25\\
GJS~\cite{gjs} & 44.4 $\pm$ 0.40 & 45.8 $\pm$ 0.67 & 49.2 $\pm$ 0.32 & 55.8 $\pm$ 0.33 & 61.6 $\pm$ 0.28 \\
NCR~\cite{ncr} & 46.0 $\pm$ 0.40 & 46.9 $\pm$ 0.51 & 50.7 $\pm$ 0.17 & 56.5 $\pm$ 0.18 & 62.4 $\pm$ 0.30 \\
Label Propagation~\cite{iscen2020graph} & 39.6 $\pm$ 0.78 & 46.5 $\pm$ 0.22 & 54.8 $\pm$ 0.42 & 59.6 $\pm$ 0.11 & 62.0 $\pm$ 0.14 \\
MLP~\cite{iscen2020graph} & 46.9 $\pm$ 0.78 & 50.1 $\pm$ 0.38 & 55.4 $\pm$ 0.29 & 59.2 $\pm$ 0.26 & 61.5 $\pm$ 0.31\\

\midrule
Iscen~\etal$^\dag$~\cite{iscen2020graph} & 48.71 $\pm$ 0.53 & 51.13 $\pm$ 0.40 & 54.26 $\pm$ 0.29 & 59.92 $\pm$ 0.34  & 63.84 $\pm$ 0.22 \\

Ours  & \textbf{49.20} $\pm$ 0.27 & \textbf{52.83} $\pm$ 0.40 & \textbf{58.60} $\pm$ 0.20 & \textbf{61.59} $\pm$ 0.28  & \textbf{64.33} $\pm$ 0.28 \\
\bottomrule
\end{tabular}

\label{table:lowshotplaces365}
\end{table*}

\textbf{Low-shot Places365}~\cite{zhou2017places} is divided into 183 test and 182 validation classes by~\cite{iscen2020graph}. Note that we treat all classes in Places365 as novel categories.

\textbf{Low-shot ImageNet}~\cite{Hariharan_2017_ICCV} is created from ImageNet. The total 1000 classes from the ImageNet dataset are divided into 389 base classes and 611 novel classes. For the purpose of cross-validation, this benchmark is split into two disjoint sets where the test set contains 196 base classes and 311 novel classes and the remaining are in the validation set.

\textbf{Noisy data statistics.} In our experiments, the above two datasets are considered as the clean sets and are both extended by large-scale noisy images from the YFCC100M dataset~\cite{thomee2016yfcc100m}. YFCC100M consists of around 100M images collected from the Flickr where each image is attached to a text description. Refer to~\cite{iscen2020graph}, noisy images are selected if their text annotations contain the name of the novel class.
Towards the end, the Low-shot Places365 and the Low-shot ImageNet datasets are supplied by extra 9,720,957 and 3,744,994 noisy images, respectively.

\subsubsection{Evaluation metric}
Following the standard evaluation protocol used in the few-shot setting~\cite{Hariharan_2017_ICCV,iscen2020graph}, we perform 5 trials under different $k \in \{1, 2, 5, 10, 20\}$ shot setup. For each trial, we sample $k$ clean images per class from the clean set and combine them with all noisy data as the training dataset. We report the average top-5 accuracy over 5-trials on the novel class in the test set.

\subsubsection{Training details}
In our experiments, features are extracted from ResNet-10 and ResNet-50 by~\cite{iscen2020graph}. The feature extractor is trained on the base class from the Low-shot ImageNet. The dimension $d$ of the input features is 512 for ResNet-10 and 256 for ResNet-50 (after PCA), respectively. For graph cleaning stage, similar to~\cite{iscen2020graph}, we use Adam with a weight decay of 0.0005 as our optimizer. The initial learning rate is set to 0.1 for 100 iterations and decays by 0.1 every 30 iterations. We divide $T = 5$ groups for all shot setups. The hyperparameters $\alpha$ and $\beta$ are tuned on the validation set. We cross-validate the possible values of $\alpha$ and $\beta$ in the interval $[0.01, 1.0]$. The step is 0.01 for $[0.01, 0.1]$ and 0.1 otherwise.
For classifier learning, we initialize the cosine classifier with the class prototype. The cosine classifier is trained with a batch size of 512 and optimized with Adam for 50 epochs. The learning rate starts from 0.1 and is finally reduced to 0.001 with cosine annealing~\cite{loshchilov2016sgdr}. We set the temperature $s = 15$. 

\begin{table*}
\centering
\small
\setlength{\tabcolsep}{3pt}
\caption{
\textbf{Comparisons between different methods on Low-shot ImageNet.} Our method outperforms other baselines consistently. Iscen \etal~$^\dag$\cite{iscen2020graph} is reimplemented by ourselves. \textbf{Best} results are highlighted.
}
\begin{tabular}{lccccc}
\toprule
\multirow{2}{*}
{\textbf{Methods}}  & \multicolumn{5}{c}{TOP-5 ACCURACY ON NOVEL CLASSES} \\
 & k=1 & k=2 & k=5 & k=10 & k=20\\ 
\midrule
\multicolumn{6}{c}{ResNet-10 - Few Clean Data} \\
\midrule
ProtoNets~\cite{NIPS2017_cb8da676} & 39.3 & 54.4& 66.3& 71.2& 73.9 \\
Class proto.~\cite{Gidaris_2018_CVPR} & 45.3 $\pm$ 0.65 & 57.1 $\pm$ 0.37 & 69.3 $\pm$ 0.32 & 74.8 $\pm$ 0.20 & 77.8 $\pm$ 0.24 \\
Class proto. w/Att.~\cite{Gidaris_2018_CVPR} & 45.8 $\pm$ 0.74 & 57.4 $\pm$ 0.38 & 69.6 $\pm$ 0.27 & 75.0 $\pm$ 0.29 & 78.2 $\pm$ 0.23 \\
\midrule
\multicolumn{6}{c}{ResNet-10 - Few Clean \& Many Noisy Data} \\
\midrule
Similarity~\cite{iscen2020graph} & 49.8 $\pm$ 0.29&56.3 $\pm$ 0.27 & 64.2 $\pm$ 0.32 & 68.4 $\pm$ 0.14 & 71.2 $\pm$ 0.12\\
$\beta$-weighting, $\beta = 1$~\cite{iscen2020graph}& 56.1 $\pm$ 0.06 & 56.4 $\pm$ 0.08 & 57.1 $\pm$ 0.05 & 57.7 $\pm$ 0.08 & 58.7 $\pm$ 0.06\\
$\beta$-weighting, $\beta^*$~\cite{iscen2020graph} & 55.6 $\pm$ 0.24 & 58.3 $\pm$ 0.14 & 63.4 $\pm$ 0.25 & 67.5 $\pm$ 0.34 & 71.0 $\pm$ 0.22\\
GJS~\cite{gjs} & 65.5 $\pm$ 0.33 & 66.9 $\pm$ 0.24 & 69.3 $\pm$ 0.32 & 73.9 $\pm$ 0.28 & 77.9 $\pm$ 0.09 \\
NCR~\cite{ncr} & 66.7 $\pm$ 0.24 & 67.8 $\pm$ 0.12 & 70.5 $\pm$ 0.20 & 74.4 $\pm$ 0.26 & 78.4 $\pm$ 0.22 \\
Label Propagation~\cite{iscen2020graph} & 62.6 $\pm$ 0.35 & 67.0 $\pm$ 0.41 & 74.6 $\pm$ 0.30 & 76.3 $\pm$ 0.23 & 77.7 $\pm$ 0.18 \\
MLP~\cite{iscen2020graph} & 63.6 $\pm$ 0.41 & 68.8 $\pm$ 0.42 & 73.7 $\pm$ 0.25 & 75.6 $\pm$ 0.21 & 77.6 $\pm$ 0.21\\

\midrule

Iscen~\etal$^\dag$~\cite{iscen2020graph} &72.88 $\pm$ 0.44&74.94 $\pm$ 0.20&76.07 $\pm$ 0.21&78.78 $\pm$ 0.25&81.02 $\pm$ 0.30 \\

Ours  &72.92 $\pm$ 0.26 &75.60 $\pm$ 0.36 &78.83 $\pm$ 0.31 &80.71 $\pm$ 0.21 &81.06 $\pm$ 0.22 \\

\midrule
\multicolumn{6}{c}{ResNet-50 - Few Clean Data} \\
\midrule
ProtoNets~\cite{NIPS2017_cb8da676} & 49.6 & 64.0 & 74.4 & 78.1 & 80.0 \\
Class proto.~\cite{Gidaris_2018_CVPR} & 50.1 $\pm$ 0.62 & 62.9 $\pm$ 0.43 & 74.9 $\pm$ 0.10 & 79.5 $\pm$ 0.25 & 82.1 $\pm$ 0.34 \\

\midrule

\multicolumn{6}{c}{ResNet-50 - Few Clean \& Many Noisy Data} \\
\midrule

GJS~\cite{gjs} & 73.0 $\pm$ 0.36 & 74.9 $\pm$ 0.35 & 77.3 $\pm$ 0.32 & 81.4 $\pm$ 0.28 & 84.6 $\pm$ 0.15 \\
NCR~\cite{ncr} & 74.7 $\pm$ 0.21 & 76.5 $\pm$ 0.14 & 79.4 $\pm$ 0.29 & 82.5 $\pm$ 0.24 & 85.4 $\pm$ 0.17 \\
\midrule
Iscen~\etal$^\dag$~\cite{iscen2020graph} & 79.57 $\pm$ 0.30 & 81.48 $\pm$ 0.32 & 81.64$\pm$0.26 & 84.75 $\pm$ 0.18 & 87.06 $\pm$ 0.14 \\
Ours & \textbf{80.30} $\pm$ 0.43 & \textbf{82.77} $\pm$ 0.24 & \textbf{85.17}$\pm$0.34 & \textbf{86.81} $\pm$ 0.10 & \textbf{87.24} $\pm$ 0.13 \\

\bottomrule

\end{tabular}
\label{table:lowshotimagenet}
\end{table*}

\subsection{Evaluation Results}
\subsubsection{Baseline Setups}
We compare our method with several baselines. These baselines include: (1) Class proto.~\cite{Gidaris_2018_CVPR}: the class prototype is computed as the mean of the clean feature embeddings. (2) ProtoNets~\cite{NIPS2017_cb8da676}: a meta-learning approach for few-shot classification. (3) $\beta$-weighting~\cite{iscen2020graph}: the relevance score is set as $\beta$. (4) Label Propagation~\cite{iscen2020graph}: the relevance scores are obtained by solving a linear system. (5) MLP~\cite{iscen2020graph}: this model learns a nonlinear mapping for assigning relevance scores. (6) Similarity~\cite{iscen2020graph}: the relevance score is calculated as the cosine similarity between the data point and the class prototype. (7) GJS~\cite{gjs}: a generalized jensen-shannon divergence loss for learning with noisy labels. (8) NCR~\cite{ncr}: a neighbor consistency loss for combating with noisy labels. (9) Iscen~\etal~\cite{iscen2020graph}: a recently proposed method based on the graph convolutional network for learning with few clean and many noisy labels. Please refer to~\cite{iscen2020graph} for more details.

\subsubsection{Quantitative analysis}
We report the top-5 accuracy performance on Low-shot Places365 in Table~\ref{table:lowshotplaces365}. $\beta$-weighting takes effect when clean images are limited. It performs worse than class proto if enough clean samples are provided. GJS and NCR get better results by constraining the network to output consistent predictions for noisy labels, but still lag behind our method by 9.4\% and 7.9\%, respectively, in the 5-shot setting. By measuring the relations between noisy and clean data, Label Propagation and MLP improve a lot in the low-shot settings. In the 20-shot setting, the improvement upon the class prototype is marginal.
We observe that our method consistently outperforms other methods, especially in the 2/5/10-shot setup. The performance gaps relative to the Iscen~\etal~are \textbf{+1.7\%}, \textbf{+4.3\%} and \textbf{+1.6\%}, respectively. In the 5-shot setting, our SimNoiPro provides 58.60\% accuracy over the Iscen~\etal~54.26\%. It indicates we can learn a better few-shot classifier with the relevance scores generated by our consistent learning objective. 

Table~\ref{table:lowshotimagenet} presents the top-5 accuracy performance of different methods on Low-shot ImageNet. First, we can find that learning with additional noisy data brings significant improvement in the few-shot setting, especially 1-shot case, where we can gain more than 20\% performance improvement. Even a simple baseline Similarity can improve 4\% accuracy after using noisy data. This confirms that noisy data are potentially useful to facilitate the learning of the few-shot classifier if we can mine those relevant images and measure their similarities with clean ones. Second, compared to other methods that also use additional noisy data to enhance the classifier learning, our method achieves better performance, mostly in the 5-shot setup, where our model can gain the average top-5 accuracy up to 78.83\% with the backbone of ResNet-10, while the Iscen~\etal~method~\cite{iscen2020graph} can only reach 76.07\%. Our SimNoiPro directly optimizes the class relevance scores for noise-tolerant prototypes generation so that it can overcome the inconsistent optimization issues in the baseline method and produce a more discriminative prototype for classification. 
Note our SimNoiPro can even offer comparable performance in the 5-shot setting to the Iscen~\etal~(78.78\%) where 10 clean images are given.
This is very beneficial and efficient when deployed to the real-world scenario because it indicates that few well-annotated clean data are good enough to identify relevant images from a noisy set and improve the generalization of the few-shot learner. We also notice that our method achieves comparable results under the $k\in\{1,20\}$ shot settings. The comparable performance on the 1-shot setting might be due to the lack of a representative clean prototype. Given only one clean image, noise cleaning can be much more biased. For the 20-shot setting, the highly discriminative capacity of the prototype composed of 20 clean images might lead to limited improvement. Last, if we use a stronger backbone such as ResNet-50 in our experiment, more discriminative features are extracted, which contributes to better performance. Our approach yields an enhancement of 3.5\% and 2.1\% in relative improvement over the strong Iscen~\etal baseline in the 5-shot and 10-shot settings, respectively.
These results suggest that our SimNoiPro works better by introducing multiple noise-tolerant prototypes to model the diversity of the noisy web image set and incorporating the class prototype generation into the noise cleaning procedure to produce more plausible relevance scores. The noise-tolerant prototype can be much more compact and discriminative with lower intra-class variance. Training with the cosine classifier can boost performance.

\subsection{Ablation Study}
In this subsection, we conduct several ablation studies: (1) classification accuracy of the class prototype; (2) different number of noise groups; (3) different types of noise-tolerant prototypes generation; (4) replace the similarity maximization loss with the similarity minimization loss.
\begin{table*}[ht]
\caption
{
\textbf{Class prototype classification.} Our method can generate more discriminative and compact prototypes for performance improvement. We report average top-5 accuracy under the 5-shot setting. Iscen~\etal$^\dag$~\cite{iscen2020graph} is reimplemented by ourselves.
}
\centering
\small
\setlength{\tabcolsep}{3pt}
\begin{tabular}{ccc} 
\toprule
\textbf{Methods} &Low-shot Places365 & Low-shot ImageNet \\
\midrule	
Iscen~\etal$^\dag$~\cite{iscen2020graph} & 53.80 $\pm$ 0.26 & 73.76 $\pm$ 0.20  \\  	
Ours & \textbf{57.16 $\pm$ 0.45}  & \textbf{75.38 $\pm$ 0.28}   \\  			
\bottomrule
\end{tabular}
\label{table:ab_proto}
\end{table*}			
\subsubsection{Classification accuracy of the class prototype}
We directly leverage the class prototype to perform the classification. The test image is classified by the
nearest matching. As seen in Table~\ref{table:ab_proto}, our method outperforms the baseline method on both Low-shot Places365 and Low-shot ImageNet. The improvement is 3.4\% and 1.6\%, respectively. Our SimNoiPro enables better relevance score generation, which results in more discriminative and compact prototypes.
\subsubsection{Effect of the number of noise-tolerant prototypes}
\begin{figure}[htbp]
  \begin{minipage}[t]{0.45\linewidth}
    \centering
    \includegraphics[width=\linewidth]{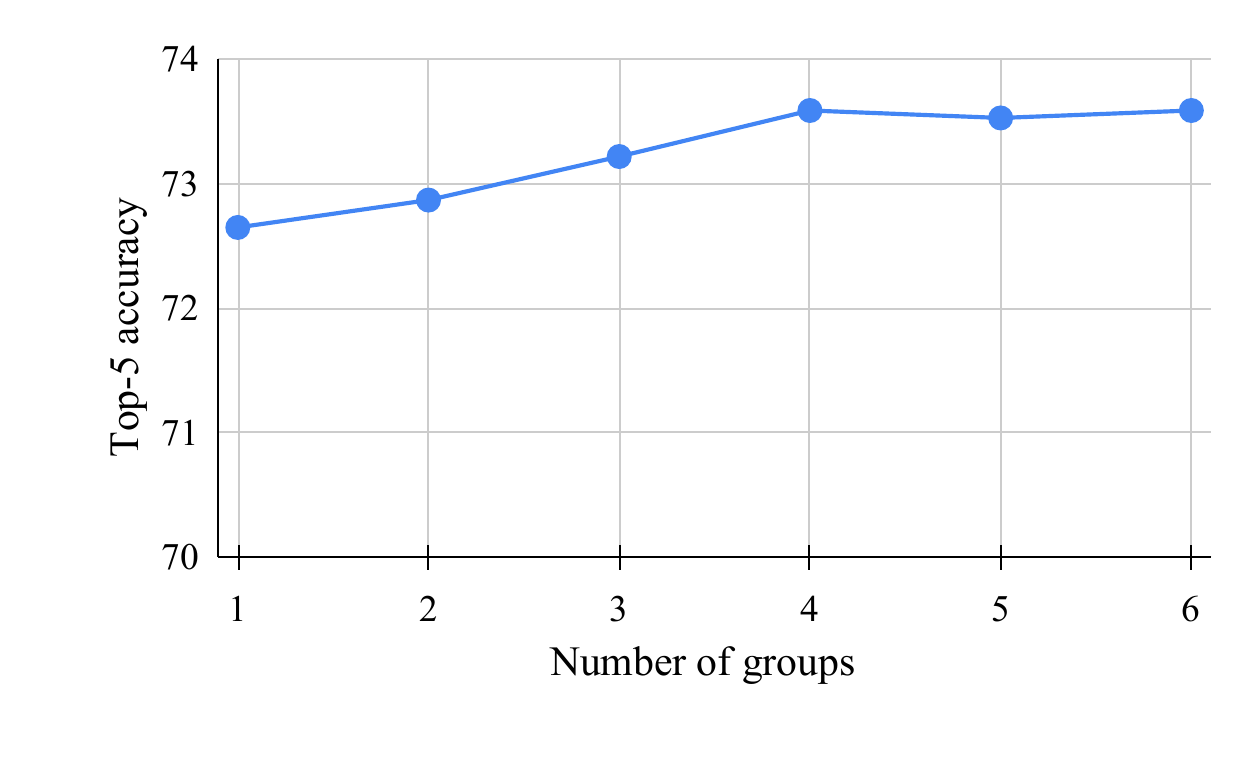}
    \caption{\textbf{Effect of the number of noise-tolerant prototypes.} More noise-tolerant prototypes can lead to better performance before saturation.}
    \label{fig:diff_groups}
  \end{minipage}%
  \hspace{3em}
  \begin{minipage}[t]{0.45\linewidth}
  \vspace{-10em}
    \centering
    \small
    \captionsetup{type=table}
     \caption{\textbf{Effect of the hyperparameter setting in our SimNoiPro loss.} Both global and local terms are important. Increasing $\alpha_t$ is preferred.}
    \begin{tabular}[t]{cc}
    \toprule
     \multirow{2}{*}
{\textbf{Methods}} & TOP-5 ACCURACY \\
     & ON NOVEL CLASSES \\
    \midrule
      $\beta$ = 0 & 66.59 $\pm$ 0.43 \\
      $\alpha_t$ = 0 & 72.68 $\pm$ 0.31 \\
      \midrule
      Decreasing $\alpha_t$ & 73.86 $\pm$ 0.28\\
      Equal $\alpha_t$ & 74.70 $\pm$ 0.22\\
      Increasing $\alpha_t$ & \textbf{75.38 $\pm$ 0.28}\\
    \bottomrule
    \end{tabular}
    \label{tab:hyperparameter_setup}
  \end{minipage}
\end{figure}

We investigate the influence of the number of noise groups on the performance in our few-shot setting. Here, we evaluate our method under the 5-shot setting.
We divide the noisy data into $T \in \{1, 2, 3, 4, 5, 6\}$ groups, and each group is assigned an equal weight, \ie, $\alpha_t$ is set to $1.0$ for all groups. The results on Low-shot ImageNet are shown in Figure~\ref{fig:diff_groups}. We observe that the top-5 classification accuracy improves when we increase the number of noise groups but reaches saturation when $T = 4$. This indicates that SimNoiPro is less sensitive to the number of noise groups when the number of noise groups is larger than 4. Compared to the case when $T=1$, more noise-tolerant prototypes bring the improvement of the performance, showing that the introduction of multiple noise-tolerant prototypes plays an important role in the modeling of the diverse noisy set.
\subsubsection{Effect of the hyperparameter setting in our SimNoiPro loss.} We study the effect of the hyperparameter by setting $\alpha_t=0$ or $\beta=0$ individually. The results are shown in Table~\ref{tab:hyperparameter_setup}. We find that removing any of them can degrade the performance. The accuracy drops by 8.8\% when $\beta=0$ and 2.7\% when $\alpha_t=0$, respectively. Besides, we perform the ablation analysis on the setup of the sequence $\{\alpha_t\}$. We compare three types: (1) Decreasing $\alpha_t$. (2) Equal $\alpha_t$. (3) Increasing $\alpha_t$. Among them, Increasing $\alpha_t$ achieves the best result. It confirms our hypothesis that a noise-tolerant prototype should be assigned a relatively more substantial proportion if it exhibits a higher degree of similarity to the clean prototype, as discussed in Section~\ref{sec:simnoisyproto}.

\subsubsection{Effect of the types of noise-tolerant prototypes generation}
In this ablation, we consider two types of noise-tolerant prototypes generation: feature based clustering and relevance score based separation. For feature based clustering, k-means is applied to cluster the pre-computed noisy features into 5 groups. We adopt similar hyperparameter configurations as the relevance score based method, and ensure that the noise-tolerant prototype closer to the clean prototype is assigned a higher weight. Table~\ref{table:ablation_separation} shows the comparison between two ways of noise-tolerant prototypes generation on Low-shot ImageNet. Clustering based method shows worse performance than the relevance score based method. When there are more clean images, the gap tends to be smaller. As discussed in Section~\ref{sec:noisy_proto_gen}, clustering is performed offline and the resulting noise groups are fixed in the graph cleaning stage. The separation is determined by the geometry of the feature space. It cannot embrace the benefit of the learned relations by graph convolutions. This leads to performance degradation. On the contrary, the relevance score based generation exhibits the advantages of adaptively producing the customized noise groups, which are much more robust to the influence of the irrelevant images. Therefore, we apply the relevance score based noise-tolerant prototypes generation in our experiments.

\subsubsection{Similarity minimization \vs Similarity maximization}
If the noisy set is pretty diverse and composed of many closely relevant images, pulling the noise-tolerant prototypes closer to the less representative clean prototype might result in overfitting issues. We investigate the effect of replacing the similarity maximization by minimization. The similarity minimization loss tries to push the noise-tolerant prototypes away from the clean prototype. Therefore, the final combined prototype is prevented from being too close to the clean prototype. We validate the similarity minimization loss with the prototype classifier under different shot settings. The top-5 accuracy results on Low-shot ImageNet are presented in Table~\ref{table:ablation_pushaway}. We observe that the performance drops a lot for each shot setting. For 2-shot case, the similarity minimization method is 52\% behind the similarity maximization method. When given more and more clean images, the gap narrows. This suggests that more irrelevant images are in the noisy set and our similarity maximum loss can well select those relevant images and assign higher relevance scores to build a more discriminative classifier. Meanwhile, our learned relevance scores can measure the clean-noisy relations better, which can help the generation of a more compact prototype.
\begin{table*}
\centering
\small
\setlength{\tabcolsep}{3pt}
\caption{
\textbf{Effect of noise-tolerant prototypes generation methods.} Relevance score based separation exhibits the
advantages of producing customized and robust groups adaptively. 
}

\begin{tabular}{lccccc}
\toprule
\multirow{2}{*}
{\textbf{Methods}}  & \multicolumn{5}{c}{TOP-5 ACCURACY ON NOVEL CLASSES} \\
 & k=1 & k=2 & k=5 & k=10 & k=20\\ 
\midrule
\multicolumn{6}{c}{ResNet-10 - Few Clean \& Many Noisy Data} \\
\midrule
Feature based clustering & 58.33 $\pm$ 0.43 & 66.27 $\pm$ 0.38 &73.32 $\pm$ 0.23 &75.87 $\pm$ 0.21 &78.26 $\pm$ 0.16 \\
Relevance score based separation & \textbf{67.64 $\pm$ 0.38} &\textbf{70.98 $\pm$ 0.25} &\textbf{75.38 $\pm$ 0.28} &\textbf{76.98 $\pm$ 0.20} &\textbf{78.39 $\pm$ 0.14}\\
\bottomrule
\end{tabular}
\label{table:ablation_separation}
\end{table*}

\begin{table*}
\centering
\small
\setlength{\tabcolsep}{3pt}
\caption{
\textbf{Comparison between similarity minimization and similarity maximization method.} Similarity maximization loss learns better clean-noisy relations that help the generation of compact prototypes.
}

\begin{tabular}{lccccc}
\toprule
\multirow{2}{*}
{\textbf{Methods}}  & \multicolumn{5}{c}{TOP-5 ACCURACY ON NOVEL CLASSES} \\
 & k=1 & k=2 & k=5 & k=10 & k=20\\ 
\midrule
\multicolumn{6}{c}{ResNet-10 - Few Clean \& Many Noisy Data} \\
\midrule
Similarity minimization & 23.38 $\pm$ 0.60& 18.91 $\pm$ 0.44& 34.28 $\pm$ 0.99& 44.31 $\pm$ 1.04& 63.74 $\pm$ 0.65\\
Similarity maximization &\textbf{67.64 $\pm$ 0.38} &\textbf{70.98 $\pm$ 0.25} &\textbf{75.38 $\pm$ 0.28} &\textbf{76.98 $\pm$ 0.20} &\textbf{78.39 $\pm$ 0.14} \\
\bottomrule
\end{tabular}
\label{table:ablation_pushaway}
\end{table*}

\begin{figure*}[t]
  \centering
  \subfloat[\label{fig:class2}]{
  \includegraphics[width=0.31\textwidth]{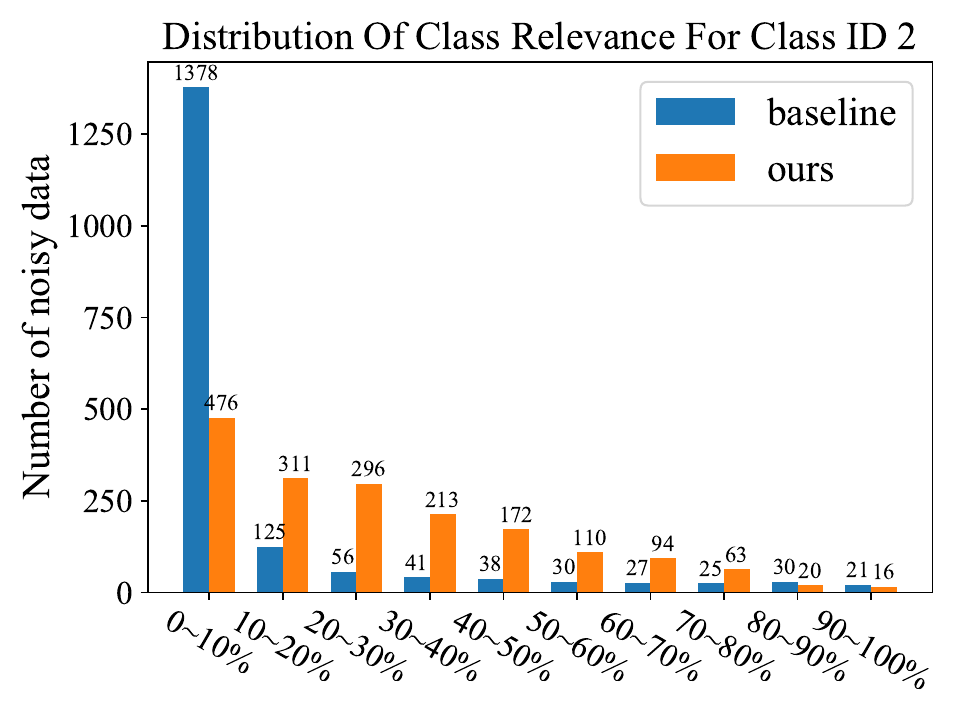} }
  \subfloat[\label{fig:class10}]{
  \includegraphics[width=0.31\textwidth]{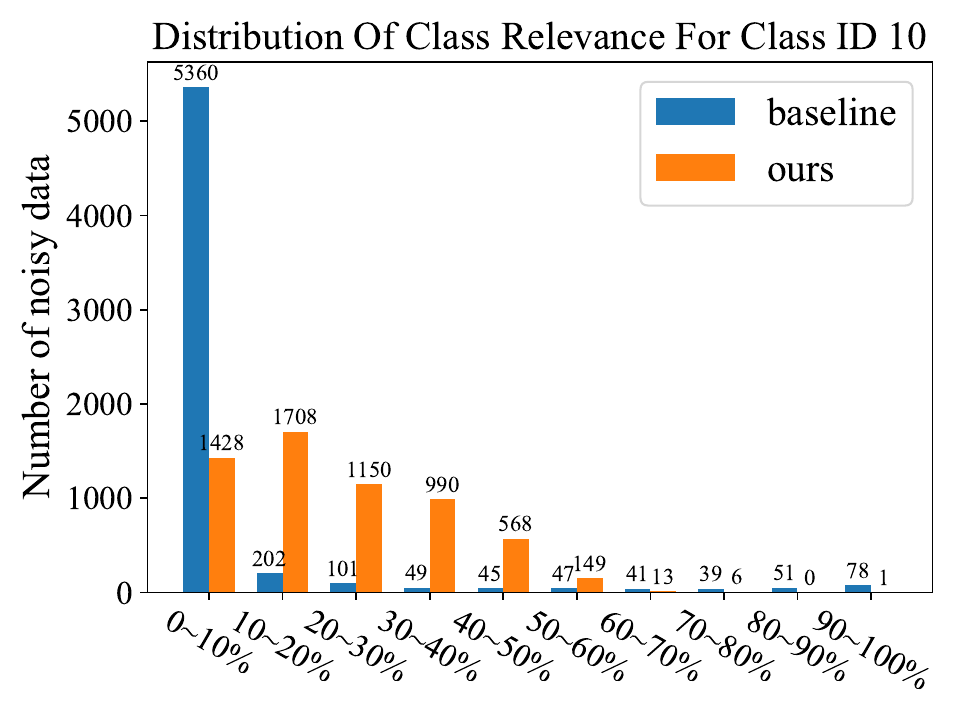}}
  \subfloat[\label{fig:class360}]{
  \includegraphics[width=0.31\textwidth]{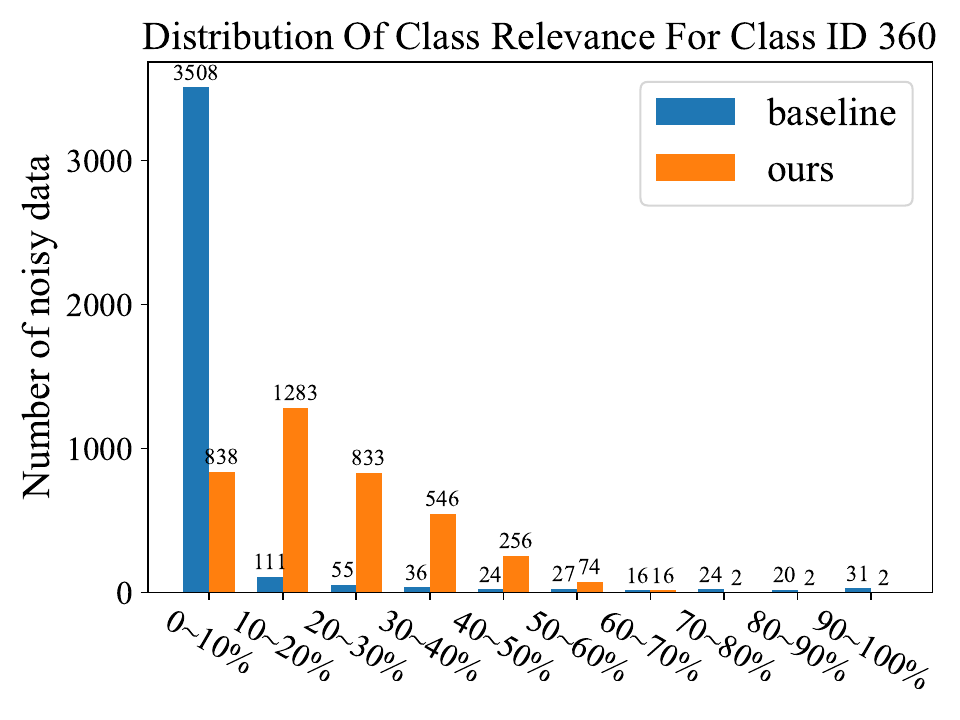}}
  \caption{
  \textbf{Class relevance $r$ distribution comparison on Low-shot Places365.} (a) is for class 2. (b) is for class 10. (c) is for class 360. Our SimNoiPro reveals the relative importance of noisy features.
  } 
\label{fig:lowshotplaces365_r}  
\end{figure*}
\begin{figure*}[t]
  \centering
  \subfloat[\label{fig:class113}]{
  \includegraphics[width=0.31\textwidth]{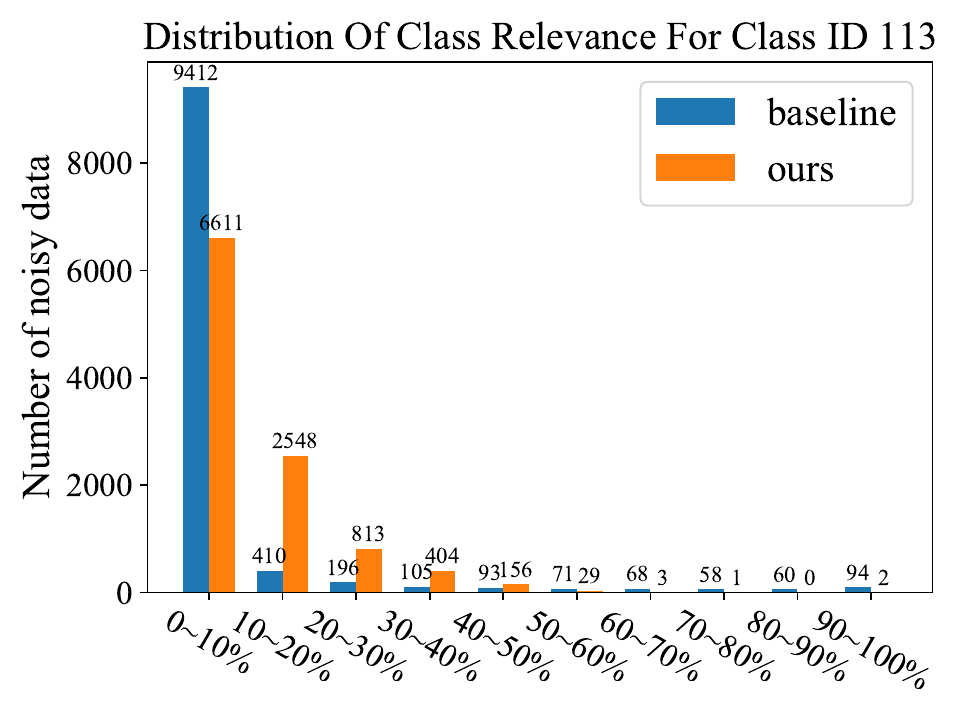}}
  \subfloat[\label{fig:class837}]{
  \includegraphics[width=0.31\textwidth]{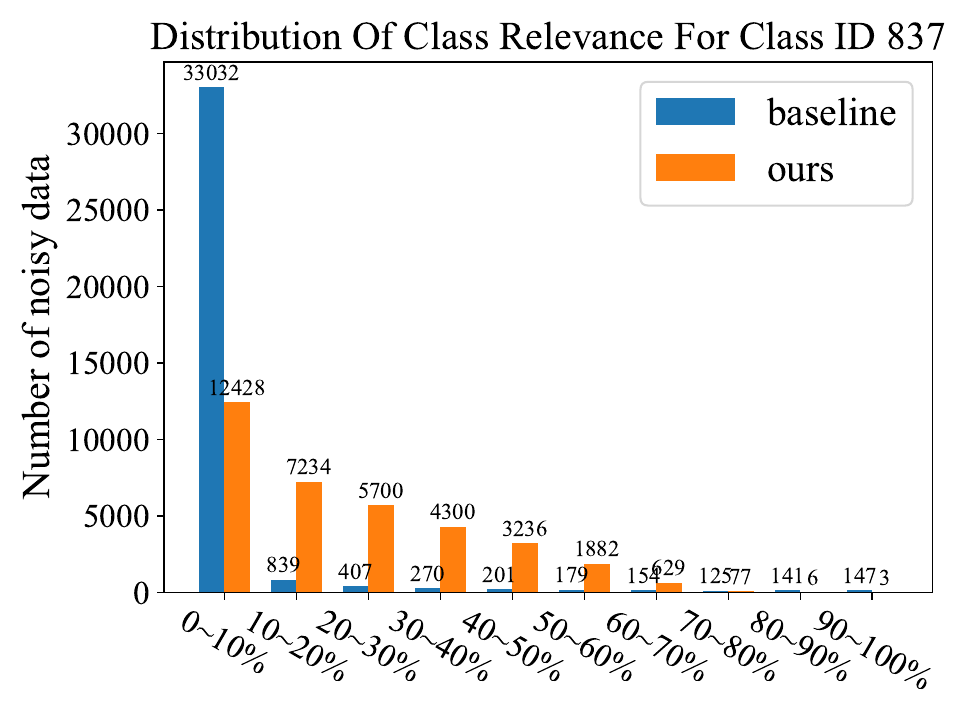}}
  \subfloat[\label{fig:class942}]{
  \includegraphics[width=0.31\textwidth]{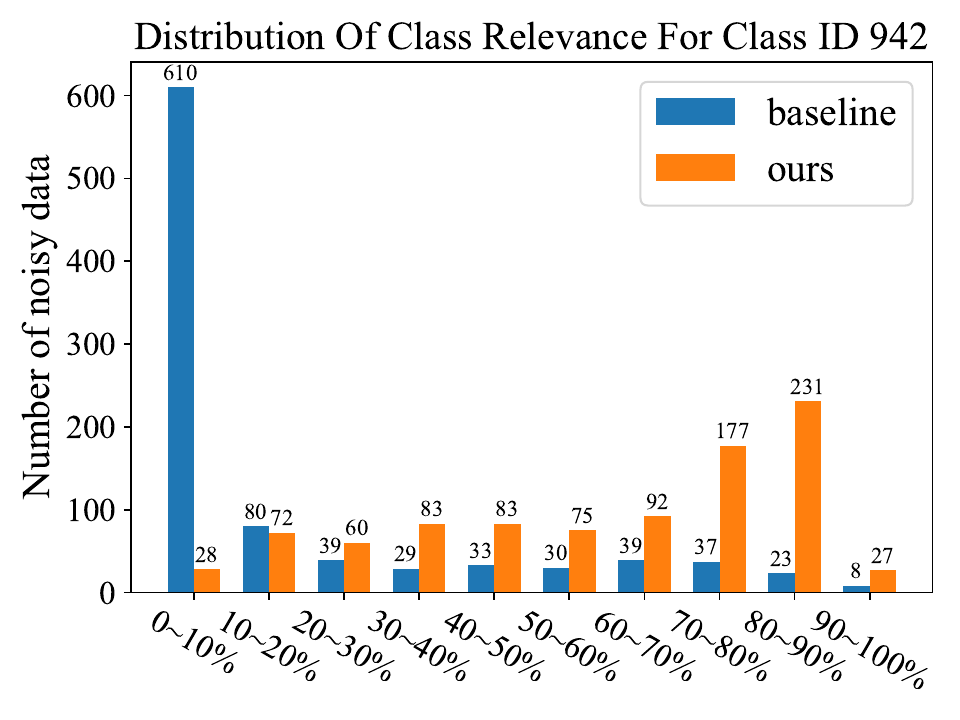}}
  \caption{
  \textbf{Class relevance $r$ distribution comparison on Low-shot ImageNet.} (a) is for class 113. (b) is for class 837. (c) is for class 942. Our SimNoiPro reveals the relative importance of noisy features.
  }
\label{fig:lowshotimagenet_r}  
\end{figure*}

\subsection{Visualization and Analysis}
In this subsection, we show several qualitative results: (1) visualization of the relevance score distribution. (2) visualization of the representative noisy images.

\subsubsection{Distribution of relevance scores}
Instead of using binary classification loss to treat the clean data as positive instances and noisy data as negative ones~\cite{iscen2020graph}, our SimNoiPro directly optimizes the relevance scores for the noise-tolerant prototypes generation in an end-to-end manner. 
We compare the distribution of the class relevance $r$ generated by our SimNoiPro and the baseline method~\cite{iscen2020graph}. In this experiment, we divide the interval $r_{max} - r_{min}$ into 10 groups equally and then count the number of noisy data for each group in the first trial under the 5-shot setting. 

The visualization results for Low-shot Places365 and Low-shot ImageNet are illustrated in Figure~\ref{fig:lowshotplaces365_r} and~\ref{fig:lowshotimagenet_r}, respectively.
In Figure~\ref{fig:lowshotplaces365_r}, we select three typical classes with higher accuracy obtained by our method on Low-shot Places365. It can be seen that for the baseline method~\cite{iscen2020graph}, most noisy data fall into the 0-10\% interval. This indicates that the binary classification loss used in~\cite{iscen2020graph} simply pushes the class relevance scores of the noisy data to be close to 0. There is no guarantee to produce a discriminative unified prototype for classification. 
However, our SimNoiPro regards each noisy data as one of the components of the noise-tolerant prototypes. 
As a result, the output of our model reveals the relative importance of noisy features and directly contributes to the unified prototype used in the following classification stage. Specifically, we observe that the distribution of $r$ produced by our method is quite different from the baseline in class 2. In Figure~\ref{fig:lowshotimagenet_r}, more visualization results on Low-shot ImageNet are presented. We also find that the distribution of the relevance scores for the baseline model is more centralized while our method produces a more diverse distribution.
\begin{figure*}[t]
\center
\includegraphics[height=0.35\linewidth]{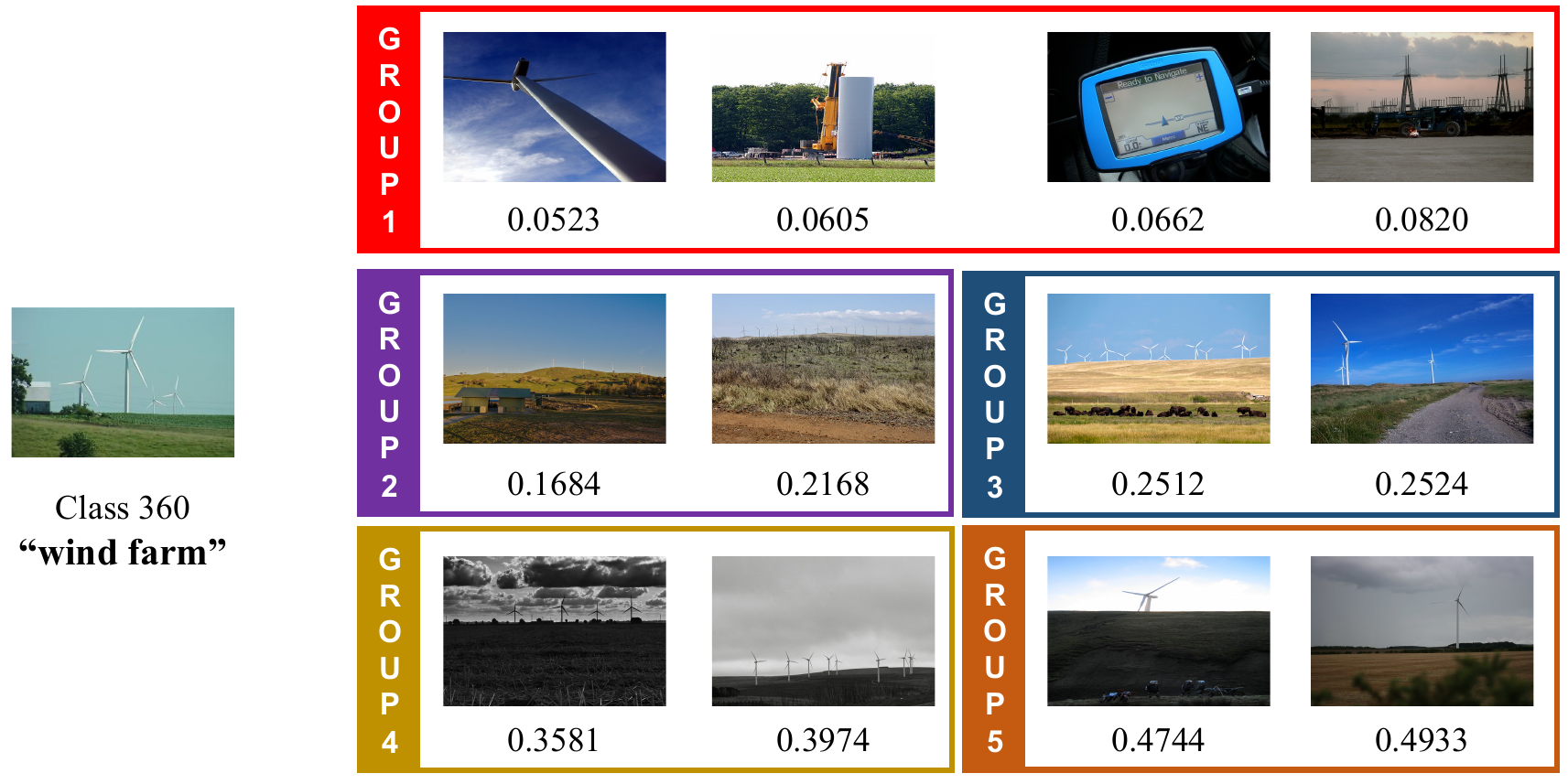}
\caption{
\textbf{Qualitative results on Low-shot Places365.} \textbf{Left} is one of the clean images with its class name ``wind farm" and class id 360. \textbf{Right} is the 5 noise groups generated by our SimNoiPro. We show each noise group mostly shares similar visual patterns. Each noisy image is annotated with its class relevance score $r$.
}
\label{fig:img_r_places365}  
\end{figure*}
\begin{figure*}[t]
\center
\includegraphics[height=0.35\linewidth]{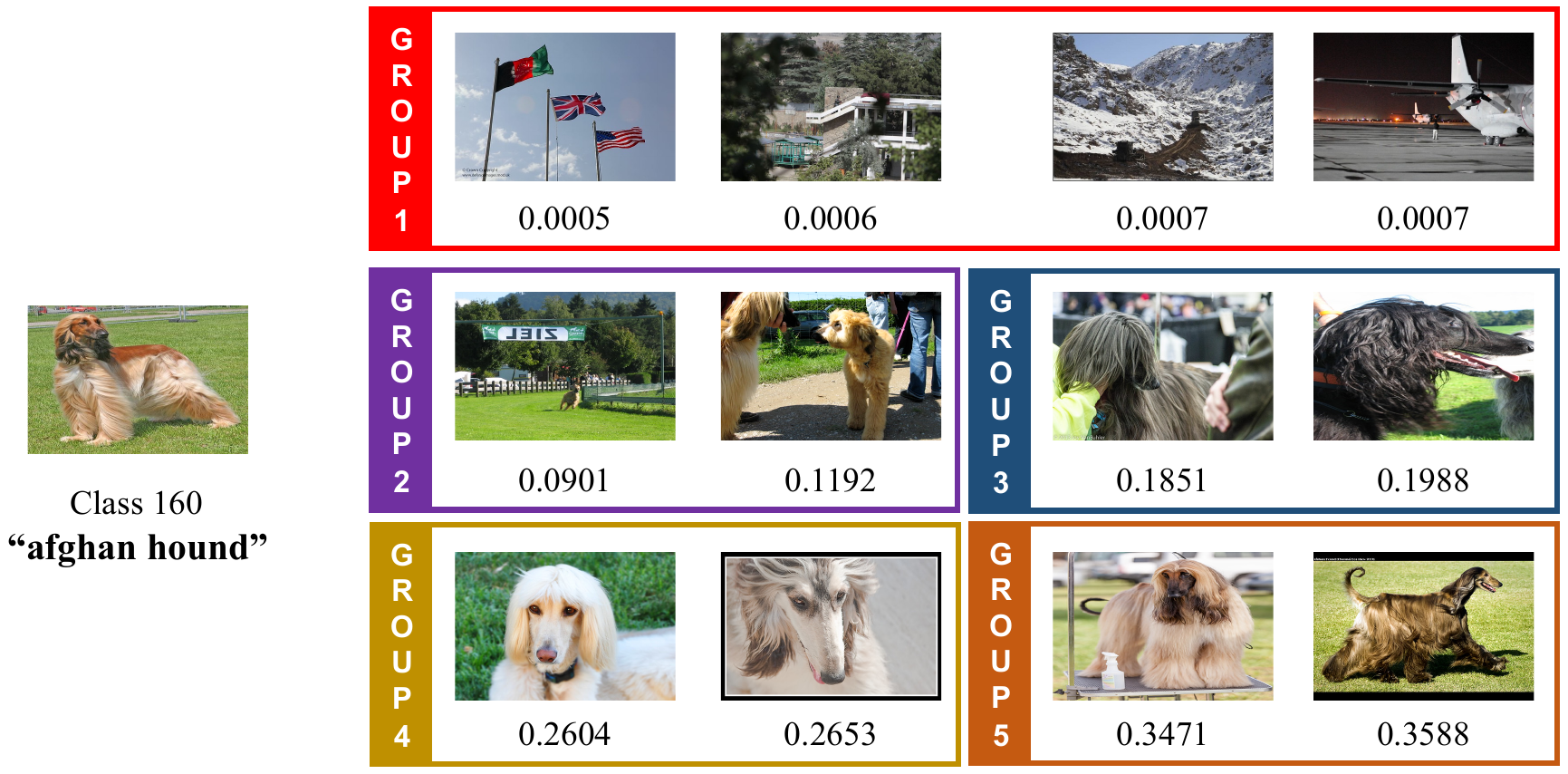}
\caption{
\textbf{Qualitative results on Low-shot ImageNet.} \textbf{Left} is one of the clean images with its class name ``afghan hound" and class id 160. \textbf{Right} is the 5 noise groups generated by our SimNoiPro. We show each noise group mostly shares similar visual patterns. Each noisy image is annotated with its class relevance score $r$.
}
\label{fig:img_r_imagenet}  
\end{figure*}

\subsubsection{Qualitative analysis}
We show the representative noisy images with the corresponding class relevance scores in the different noise groups produced by our SimNoiPro. The results on Low-shot Places365 and Low-shot ImageNet are presented in
Figure~\ref{fig:img_r_places365} and~\ref{fig:img_r_imagenet}. Figure~\ref{fig:img_r_places365} depicts the results for class ``wind farm'' on Low-shot Places365. In Figure~\ref{fig:img_r_imagenet}, we present the noise groups for class ``afghan hound''. Here, noise groups are divided by the class relevance scores. Note that our relevance score can represent the relative importance. For each noise group, we randomly sample several representative noisy images. We notice that each group mostly shares similar visual semantics except in noise group 1 which contains many irrelevant images. Our method can assign higher scores to those that look very similar to the given clean image. These results support our basic motivation in Section~\ref{sec:noisy_proto_gen}. As the noisy set is pretty diverse, using one coarse noise-tolerant prototype can fail to represent the complex noisy data collections. Our method can well model the noise data by introducing multiple prototypes. The generated relevance scores are more plausible.

\section{Conclusion}
\label{sec:conclusion}

In this paper, we introduce SimNoiPro, a similarity maximization loss, to learn a robust few-shot classifier by leveraging large-scale noisy web data. Our approach is different from previous methods that formulate noise data cleaning as a binary classification problem, which ignores the diverse nature of noisy web images and can lead to a discrepancy issue when applied to prototype-based classification. SimNoiPro introduces noise-tolerant hybrid prototypes to provide finer modeling of the diverse noisy set. It enables end-to-end learning by pulling the noise-tolerant prototypes closer to the clean prototype. We extensively evaluate SimNoiPro on Low-shot Places365 and Low-shot ImageNet, and demonstrate that it outperforms other methods, showcasing its effectiveness.


\bibliographystyle{ACM-Reference-Format}
\bibliography{sample-base}

\end{document}